\documentclass[runningheads]{llncs}
\usepackage[T1]{fontenc}
\usepackage{graphicx}
\usepackage{amsmath}
\usepackage{booktabs}
\usepackage{float}
\usepackage{subcaption}
\usepackage{comment}
\usepackage{hyperref}
\begin{document}
\title{Markerless Pose Estimation for Resistance Training Technique Assessment}
\titlerunning{Pose Estimation for Resistance Training}

\author{Joseph Turner \and
Jeff Clark\orcidID{0000-0003-0118-3999} \and
Nawid Keshtmand\orcidID{0009-0008-5552-1395}}

\authorrunning{J. Turner et al.}

\institute{School of Engineering Mathematics and Technology, University of Bristol, UK
\email{nawid.keshtmand@bristol.ac.uk}}

\maketitle              
\begin{abstract}
\label{sec:Abstract}
Resistance training can be a high risk activity, and safe form is essential to avoiding injury. Laboratory-based movement analysis provides quantitative technique assessment, yet is not easily accessible. Markerless pose estimation infers body landmarks from images or video without physical markers and could offer a feasible alternative for technique assessment. We present a pose estimation framework to evaluate resistance-training technique from ordinary video footage. Using BlazePose \cite{Bazarevsky2020BlazePose}, anatomical landmarks were extracted from squat, bench press, and deadlift videos and converted into joint-angle trajectories, with the squat serving as the primary case study. Trajectories were assessed against a defined reference repetition using root mean square error (RMSE). Results show that the framework recovers meaningful kinematic patterns for the squat and deadlift, enabling quantitative comparison between repetitions and identification of technique variability within a set. Performance depended strongly on camera orientation and visual occlusion, with non-sagittal views distorting 2D joint-angle estimates. The findings demonstrate that markerless pose estimation can support accessible biomechanical assessment outside laboratory environments.


\end{abstract}
\section{Introduction}
\label{sec:Intro}
Efficient and safe movement is central to sports performance, rehabilitation, and injury prevention \cite{bahr2005understanding}. In resistance training, technique is typically assessed through subjective coaching or, less commonly, laboratory-based biomechanical analysis \cite{MERCADALBAUDART2024e27596}. Marker-based motion capture systems such as Vicon provide highly accurate kinematic measurements but require expensive specialised equipment and controlled environments, limiting their use in everyday training settings \cite{Barris2008Vision,vicon2026}.

Recent advances in computer vision have made markerless pose estimation a promising alternative. Models such as BlazePose infer anatomical landmarks directly from ordinary video, enabling extraction of kinematic information from smartphone recordings at low cost \cite{Bazarevsky2020BlazePose,Ultralytics2024PoseEstimation}. This creates opportunities for accessible movement analysis in training and rehabilitation \cite{souaifi2025artificial}.

This paper aims to determine whether markerless pose estimation can provide a practical and interpretable basis for biomechanical assessment of resistance-training exercises outside laboratory environments. To investigate this, we developed and evaluated a framework for analysing resistance-training technique from standard video footage using pose estimation. Anatomical landmarks were extracted from recordings of the squat, bench press, and deadlift and converted into joint-angle trajectories. The squat served as the primary case study, with knee and trunk angles used for detailed biomechanical analysis, and quantitatively compared against gold standard reference repetitions. This work is intended as a feasibility study demonstrating the potential of markerless pose estimation for accessible resistance-training assessment rather than a fully validated biomechanical assessment system.

\noindent The main contributions of this paper are:
\begin{itemize}
\item Development and evaluation of a pose-estimation framework for analysing resistance-training technique from standard video footage
\item Curation of a resistance-training dataset, including reference repetitions demonstrating good technique
\item Investigation of the effects of camera viewpoint on joint-angle estimation
\end{itemize}

\section{Related Work}
\label{sec:Related Work}

Markerless approaches estimate anatomical landmarks directly from video, offering an accessible solution for coaching, rehabilitation, and performance monitoring \cite{Colyer2018}. Their accuracy, however, remains sensitive to camera viewpoint, occlusion, and image quality \cite{Wade2022}.

Human pose estimation has progressed from part-based models \cite{yang2011articulated} to deep learning approaches such as HRNet \cite{sun2019deep}, VideoPose3D \cite{pavllo20193d}, and transformer-based models such as ViTPose \cite{Xu2022ViTPose}. These methods have significantly improved landmark detection and, in some cases, enabled monocular 3D reconstruction. For practical applications, lightweight models such as BlazePose \cite{Bazarevsky2020BlazePose} are particularly attractive because they provide real-time full-body tracking from standard video.

For sports applications, OpenPose has been applied to elite long-jump competition footage, demonstrating that meaningful biomechanical measures can be extracted outside laboratory conditions \cite{Cronin2024OpenPoseCompetition}. More recently, AthletePose3D introduced a large-scale dataset of athletic movements to improve model performance on high-speed sports actions \cite{wang2025athletepose3d}.

In resistance training, computer vision has been used to estimate external variables such as barbell velocity \cite{aagren2024enhancing}. However, assessing movement quality requires joint-level biomechanical analysis. Squat biomechanics studies have shown that knee and trunk motion are key determinants of loading and technique \cite{Escamilla2001}. Mercadal-Baudart et al. demonstrated that joint angles derived from single-camera pose estimation can provide interpretable metrics for strength exercises \cite{MERCADALBAUDART2024e27596}.
Despite these advances, limited work has evaluated how robustly markerless pose estimation can assess complex powerlifting movements under realistic gym conditions. In particular, the effects of camera viewpoint, occlusion, and intra-set variability remain under explored, which this paper seeks to address.

\section{Method}
\label{sec:method}

\subsection{Dataset Curation and Preprocessing}

Two publicly available Kaggle datasets~\cite{roboflow_deadlift_dataset,kaggle_gym_workout_dataset} were combined (n = 2618 videos). The three target powerlifts (squat, bench press, and deadlift) were selected to focus on compound movements involving multiple joints and a higher risk of injury. As the original datasets contained a wide variety of gym exercises, the majority of videos were excluded during this initial selection step. The remaining recordings were then filtered to retain only videos suitable for reliable biomechanical analysis. Clips with poor lighting, severe occlusion, incomplete body visibility, or extreme camera viewpoints were removed, as these conditions prevent consistent landmark detection and can introduce substantial errors into 2D joint-angle estimation. Following this filtering process, the final dataset comprised 89 videos, which were subsequently normalised to a consistent frame rate and resolution before analysis.

To enable quantitative comparison between repetitions, a reference trajectory was established from a curated second set of instructional videos sourced from publicly available material, selected for technical soundness against widely accepted coaching standards. 
Each repetition was then evaluated against four biomechanical criteria. For the squat these were: (i) sufficient depth, indicated by a low minimum knee angle; (ii) a smooth, continuous U-shaped descent-ascent profile; (iii) stable trunk angle throughout; and (iv) minimal discontinuities attributable to pose estimation noise. The repetition best satisfying all four criteria was retained as the reference trajectory for subsequent comparisons. Reference repetitions could be tailored to an individual, providing quantified feedback tailored to the athlete's specific anatomy and training style.

\subsection{Pose-Based Biomechanical Analysis Framework}

BlazePose is a lightweight convolutional neural network designed for real-time 2D human pose estimation~\cite{Bazarevsky2020BlazePose}. This was implemented as the foundation of the analysis pipeline. For each frame $t$, the model returns a set of $M = 33$ anatomical landmarks
\begin{equation}
    \mathbf{p}_i^{(t)} = \bigl(x_i^{(t)},\, y_i^{(t)}\bigr), \quad i = 1, \ldots, M,
    \label{eq:landmarks}
\end{equation}
where $\bigl(x_i^{(t)}, y_i^{(t)}\bigr)$ are the 2D image-plane coordinates of the $i$-th keypoint. Joint angles were derived from landmark triplets using a standard vector formulation: for three points $A$, $B$, $C$ where $B$ is the joint of interest,
\begin{equation}
    \theta = \cos^{-1}\!\left(\frac{(A - B)\cdot(C - B)}{\|A - B\|\,\|C - B\|}\right).
    \label{eq:joint_angle}
\end{equation}
For the squat, the knee angle was computed from the hip-knee-ankle triplet, and trunk posture was quantified as the angular deviation of the shoulder-hip segment from the vertical axis. For the deadlift, trunk angle was derived using the same formulation, with a hip-hinge angle defined from the shoulder-hip-knee triplet to capture the degree of forward flexion at the hip. For the bench press, upper-limb kinematics were quantified using the shoulder--elbow--wrist triplet to compute elbow flexion angle, defined as the internal angle at the elbow joint.

\subsection{Temporal Normalisation}
\label{sec:temporal}

Since repetitions differ in duration due to differences in pacing and frame count, direct frame-by-frame comparison is not meaningful without temporal alignment. Linear interpolation was applied to resample each joint-angle signal onto a common normalised time axis. For a repetition of $T$ frames, the normalised coordinate was defined as:
\begin{equation}
    \hat{\tau}_t = \frac{t}{T-1}, \quad t = 0, 1, \dots, T-1
    \label{eq:temporal_norm}
\end{equation}

\noindent Each joint-angle signal was then linearly interpolated and resampled onto a common normalised time axis
\begin{equation}
\tau_i = \frac{i}{100}, \quad i = 0, 1, \dots, 100,
\end{equation}
yielding 101 samples per repetition regardless of its original frame count.
This process preserves the shape of each movement trajectory whilst removing variability due to execution speed. The reference was normalised using the same procedure, ensuring all comparisons were made on a common movement axis.

For multi-repetition analysis, a phase-aware variant was also applied. Since a multi-repetition recording contains several consecutive movement cycles, simple linear normalisation cannot distinguish between individual repetitions or align their phases meaningfully. Each repetition was therefore first segmented, then divided into descent, hold, and ascent phases, assigned fixed proportions of 40\%, 20\%, and 40\% of the full movement cycle respectively, with each phase independently normalised to [0, 1] prior to concatenation. The 40--20--40 proportions were selected as a simple heuristic to preserve the approximate temporal structure of a squat repetition, where descent and ascent typically occupy most of the movement and any pause at the bottom is comparatively brief. More adaptive phase alignment approaches, such as dynamic time warping, represent an important direction for future work.

\subsection{Joint Angle Trajectory Comparison}
\label{sec:rmse}

Following reference repetition trajectory selection, deviation between each test repetition and the reference trajectory was quantified using RMSE,
which was computed independently for the knee-angle and trunk-angle trajectories.
To express RMSE on a more interpretable scale, each value was converted to a similarity score ranging from 0 to 100:
\begin{equation}
    S = \max\!\left(0,\ 100 - 2\,\text{RMSE}\right).
    \label{eq:similarity}
\end{equation}
The scaling factor of two was selected based on the typical RMSE values observed in the dataset, allowing the resulting scores to span an interpretable 0--100 range. The similarity score is not intended as a universally validated measure of exercise technique quality, but rather as a task-specific metric for quantifying consistency between a test repetition and a reference trajectory. This provides a simple and interpretable measure of movement agreement suitable for automated assessment using markerless pose estimation. An overall similarity score is reported as the mean of the knee and trunk scores.

\subsection{Intra-Set Variability Analysis}

Repetitions were segmented by detecting local minima of the knee-angle trajectory, each corresponding to the bottom position of a squat. Only complete descent-ascent cycles were retained for analysis. Each segmented repetition was independently time-normalised 
and per-repetition metrics were computed.
Set-level summary statistics (mean RMSE, standard deviation, and best and worst repetitions by RMSE) were used to quantify intra-set technical consistency. The pipeline's code will be released upon publication.

\newpage
\section{Results and Discussion}

\subsection{Pose Estimation Performance Across Exercises}

The pipeline was evaluated on the dataset of squat, deadlift, and bench press 
recordings. Squats and deadlifts performed reliably, achieving mean usable-frame 
rates of 99.7\% and 99.0\% respectively. Bench press was substantially less 
reliable, with only 73.6\% usable frames and 77.2\% of videos successfully 
processed, reflecting the impact of barbell occlusion and supine body orientation 
on landmark extraction. These results indicate that markerless tracking performs best for upright 
exercises where the limbs of interest remain visible throughout the movement 
cycle. Consequently, the remaining analysis focuses primarily on squat and 
deadlift movements.

\begin{figure}[htbp]
    \centering
    \begin{subfigure}{0.32\linewidth}
        \includegraphics[width=\linewidth]{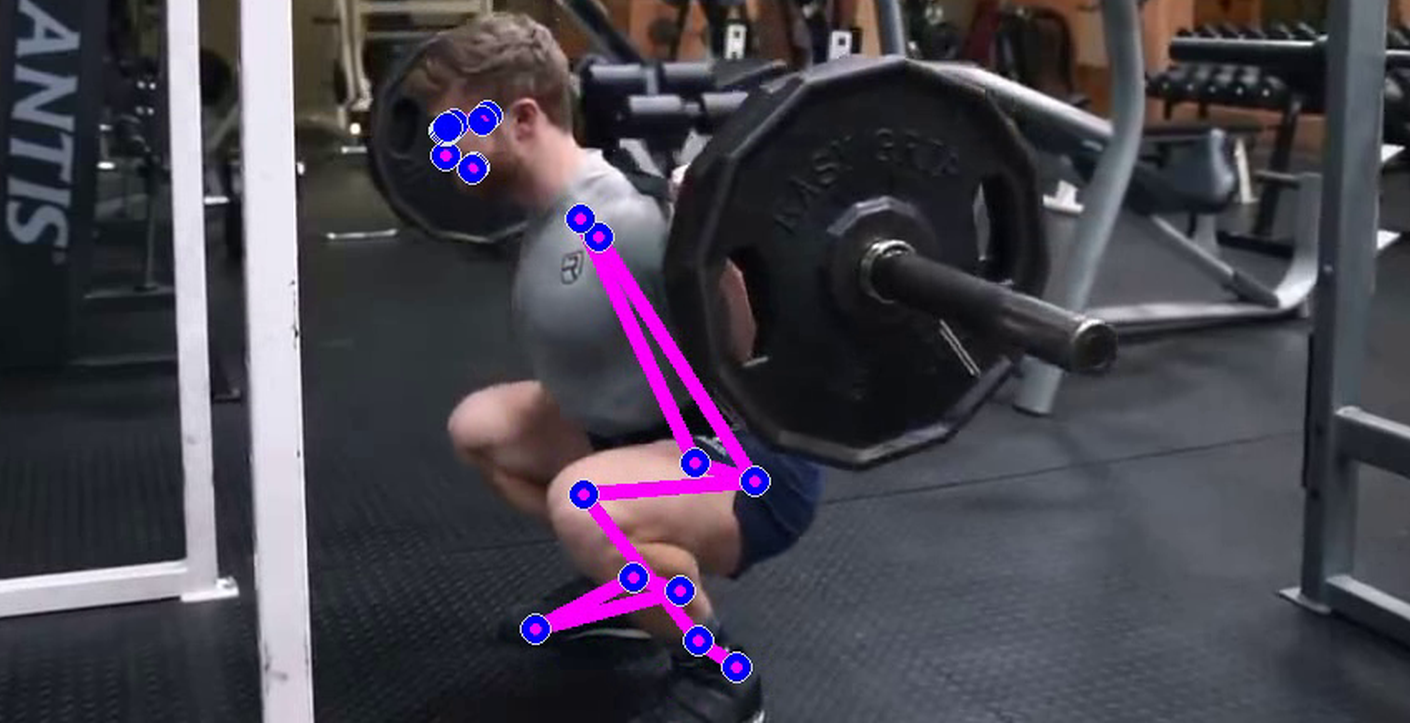}
        \caption{Squat}
        \label{fig:pose_squat}
    \end{subfigure}\hfill
    \begin{subfigure}{0.32\linewidth}
        \includegraphics[width=\linewidth]{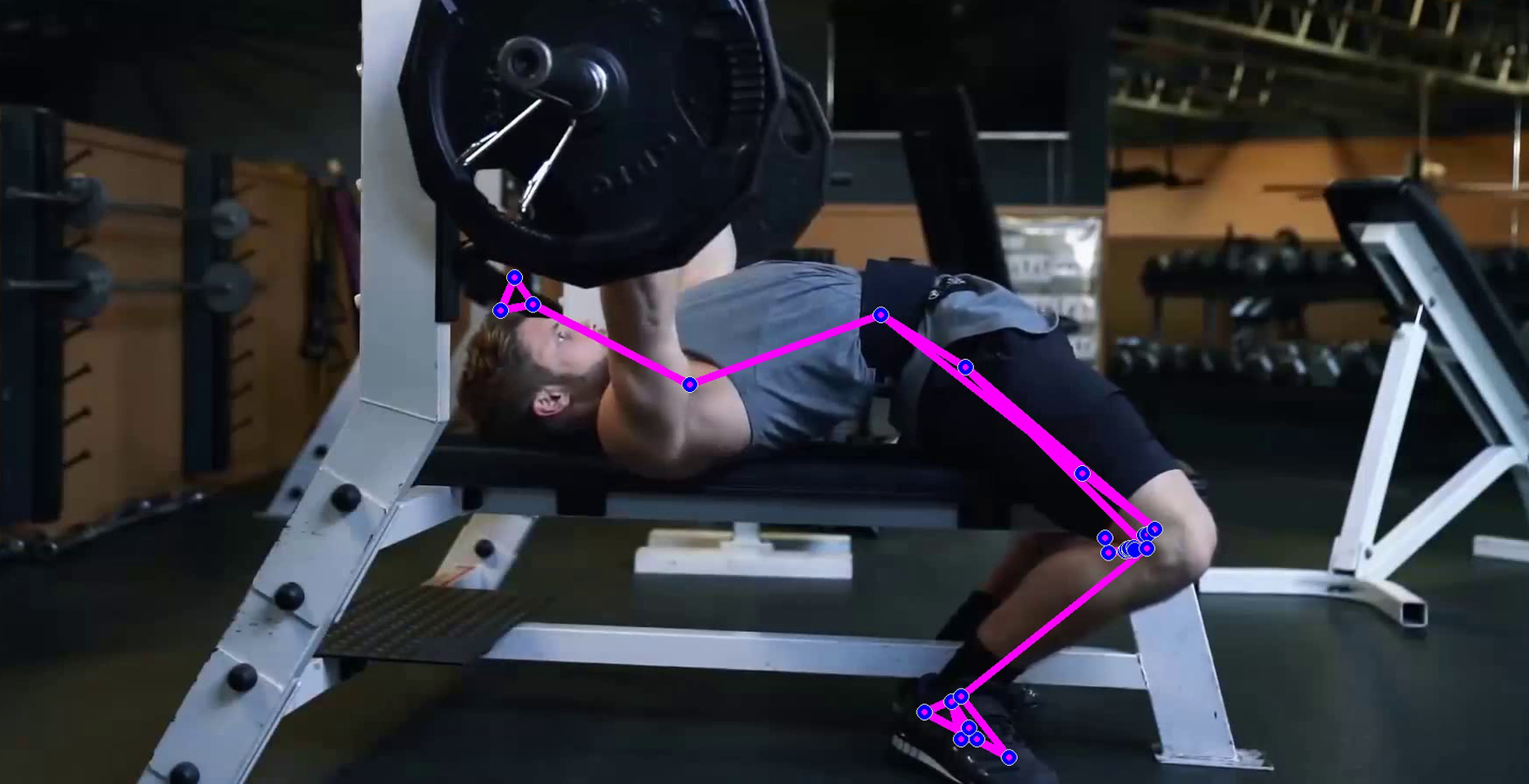}
        \caption{Bench press}
        \label{fig:pose_bench}
    \end{subfigure}\hfill
    \begin{subfigure}{0.32\linewidth}
        \includegraphics[width=\linewidth]{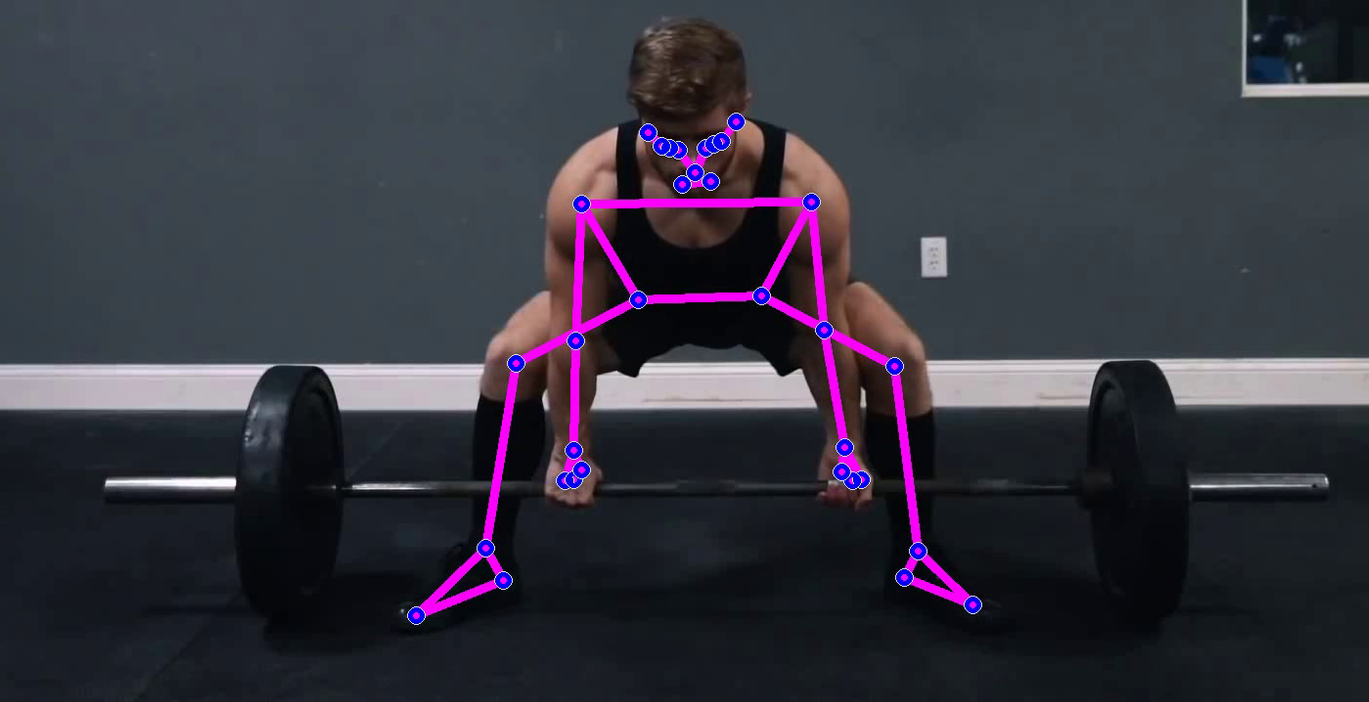}
        \caption{Deadlift}
        \label{fig:pose_deadlift}
    \end{subfigure}
    \caption{Representative examples of markerless pose estimation across the 
    three lifts. (a) Squat: the upright side-on view provides clear lower-limb visibility. (b) Bench press: upper-limb landmarks (i.e. shoulder, elbow, wrist) are difficult to track due to 
        supine position and barbell occlusion during the repetition. (c) Deadlift: the upright setup supports consistent tracking of 
        main joint landmarks.
    }
    \label{fig:pose_estimation}
\end{figure}

\subsection{Squat Trajectories and Reference Trajectory Comparison}

\begin{figure}[htbp]
\centering
\includegraphics[width=0.95\linewidth]{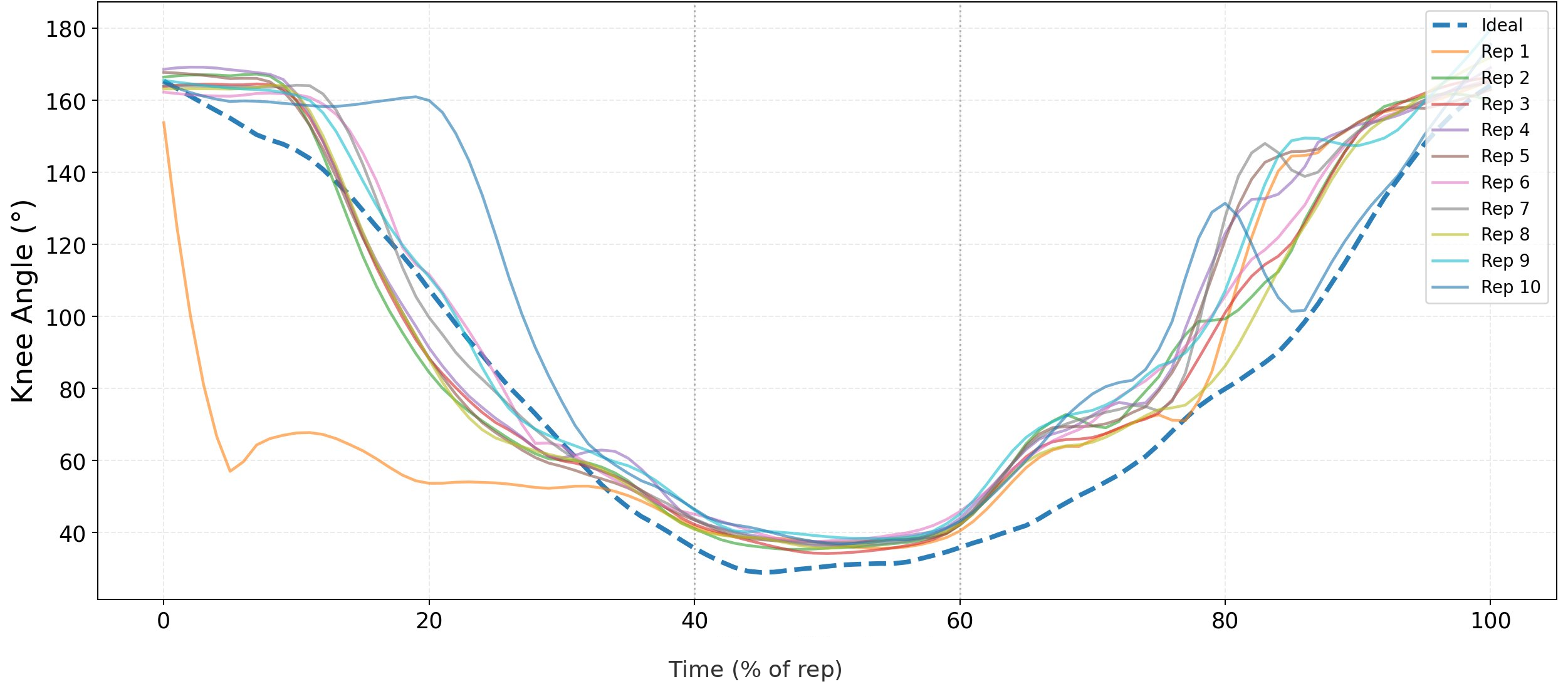}
\caption{Time-normalised knee trajectories for all ten repetitions within a squat set compared with the reference repetition trajectory (dashed line).}
\label{fig:multi_knee}
\end{figure}

Joint-angle trajectories extracted from squat videos demonstrated the expected biomechanical structure. When applied to multi-repetition analysis for a ten-repetition squat set, after temporal normalisation, most repetitions exhibited the characteristic U-shaped knee profile associated with squat descent and ascent, with minimum knee angles ranging approximately between $25^\circ$ and $50^\circ$ (Figure~\ref{fig:multi_knee}).

Most repetitions followed similar movement profiles and remained close to the reference repetition trajectory (dashed blue line, Figure~\ref{fig:multi_knee}). Quantitative results are reported in Table~\ref{tab:intraset_metrics}, with a mean RMSE of $17.56^\circ$ and mean similarity score of 64.9/100. Rep~1 showed the largest deviation from the reference repetition trajectory (RMSE $37.50^\circ$), whereas Rep~8 produced the closest agreement (RMSE $13.18^\circ$). These findings demonstrate that the framework can identify technique variability both within individual repetitions and across an entire set.

\begin{table}[htbp]
\centering
\begin{tabular}{lcccc}
\toprule
Rep & Knee Min ($^\circ$) & Knee Max ($^\circ$) & Knee RMSE ($^\circ$) & Knee Similarity \\
\midrule
1  & 35.5 & 173.9 & 37.50 & 25.0/100 \\
2  & 35.4 & 166.9 & 15.62 & 68.8/100 \\
3  & 34.1 & 168.3 & 14.15 & 71.7/100 \\
4  & 36.3 & 169.9 & 19.24 & 61.5/100 \\
5  & 37.3 & 168.1 & 19.80 & 60.4/100 \\
6  & 37.6 & 165.2 & 15.61 & 68.8/100 \\
7  & 36.4 & 165.1 & 19.45 & 61.1/100 \\
8  & 34.8 & 172.3 & 13.18 & 73.6/100 \\
9  & 38.0 & 176.7 & 18.74 & 62.5/100 \\
10 & 36.9 & 179.3 & 22.33 & 55.3/100 \\
\midrule
Mean & & & 17.56 & 64.9/100 \\
SD   & & & 6.56 & \\
\bottomrule
\end{tabular}
\caption{Per-repetition knee-angle metrics for the analysed squat set using phase-aware normalisation. RMSE is computed relative to the reference repetition trajectory.}
\label{tab:intraset_metrics}
\end{table}

\subsection{Camera Sensitivity and Occlusion Effects}
\label{sec:camera sens}
\begin{figure}[htbp]
    \centering
    \includegraphics[width=\textwidth]{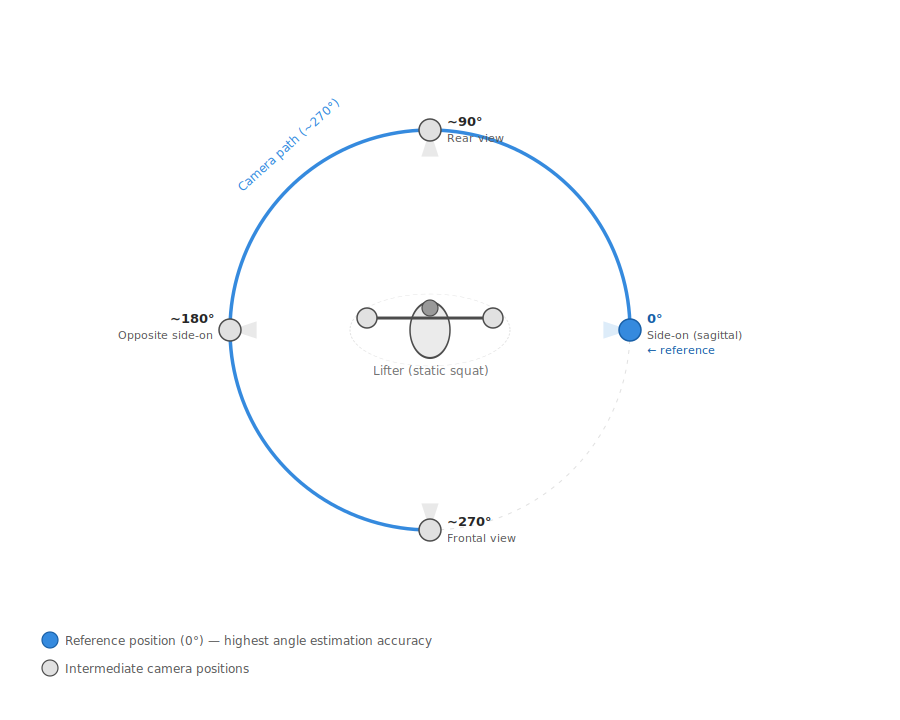}
    \caption{Bird's eye view of the camera rotation experiment. The lifter maintained a static squat posture at the centre whilst the camera was moved approximately $270^\circ$ around the subject. The blue circle marks the sagittal side-on position ($0^\circ$), which served as the reference angle due to its superior joint angle estimation accuracy. Key intermediate viewpoints rear (${\sim}90^\circ$), opposite side-on (${\sim}180^\circ$), and frontal (${\sim}270^\circ$) are also marked to illustrate the range of camera orientations evaluated.}
    \label{fig:camera_rotation}
\end{figure}

One of the strongest findings of this work was the sensitivity of joint-angle estimation to viewpoint, seen in the camera-rotation experiment (Figure~\ref{fig:camera_rotation}).

\begin{figure}[htbp]
\centering
\includegraphics[width=\linewidth]{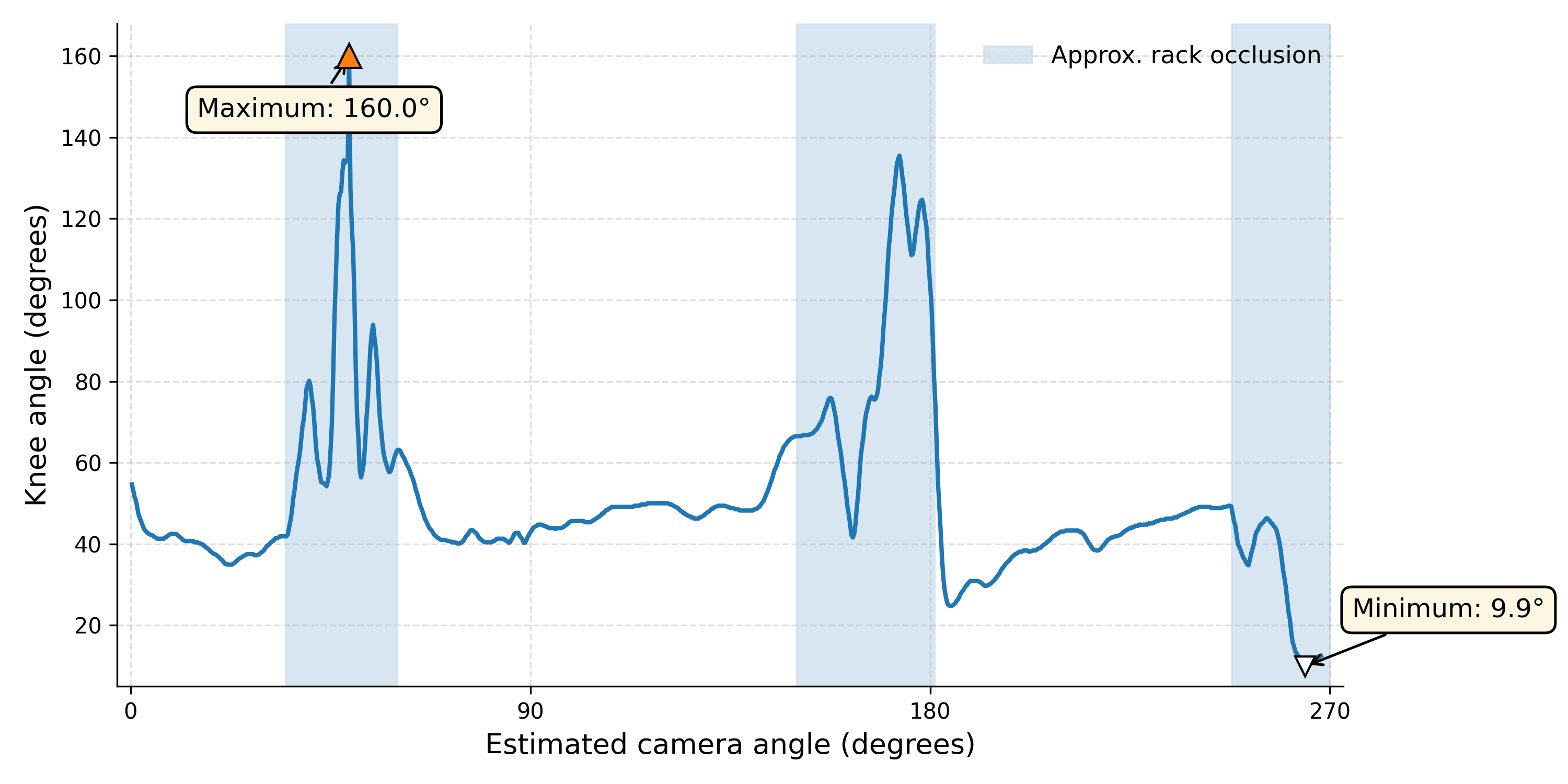}
\caption{Estimated knee angle during camera rotation around a static squat posture inside a power rack. Occlusion regions (shaded blue) produce larger deviations from the side-on reference estimate. See Figure~\ref{fig:camera_rotation} for details on the experimental setup and corresponding camera angles.}
\label{fig:rotation_rack}
\end{figure}

It showed that estimated knee angle changed substantially even when the participant maintained a fixed squat posture. The effect became most pronounced inside a power rack where lower-limb landmarks became partially obscured~(Figure~\ref{fig:rotation_rack}). The largest deviations occur when the rack partially blocks the lower limbs. Consequently, apparent changes in knee angle may reflect projection distortion rather than genuine biomechanical differences, a limitation consistent with the occlusion and viewpoint sensitivity reported by Wade et al \cite{Wade2022}. This explains several examples in the dataset where visually acceptable squats failed to produce the expected trajectory shape. The results therefore identify sagittal recording as the preferred viewpoint for practical deployment.

\subsection{Joint-Angle Trajectories for a Representative Deadlift Trial}

To demonstrate that the proposed framework generalises beyond the squat, a representative deadlift recording was analysed from a sagittal viewpoint. The same temporal normalisation procedure described for the squat was applied prior to trajectory comparison.

\begin{figure}[htbp]
    \centering

    \begin{subfigure}{0.8\linewidth}
        \centering
        \includegraphics[width=\linewidth]{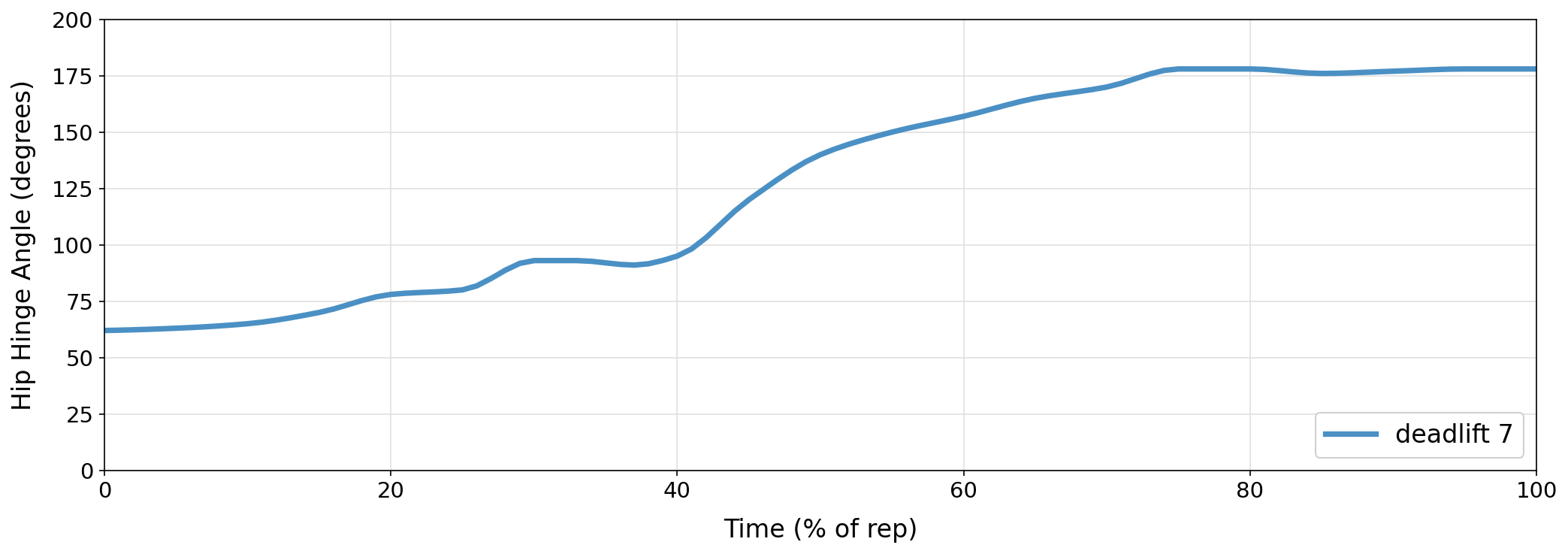}
        \caption{Time-normalised hip hinge angle.}
        \label{fig:deadlift_hip}
    \end{subfigure}

    \vspace{0.5em}

    \begin{subfigure}{0.8\linewidth}
        \centering
        \includegraphics[width=\linewidth]{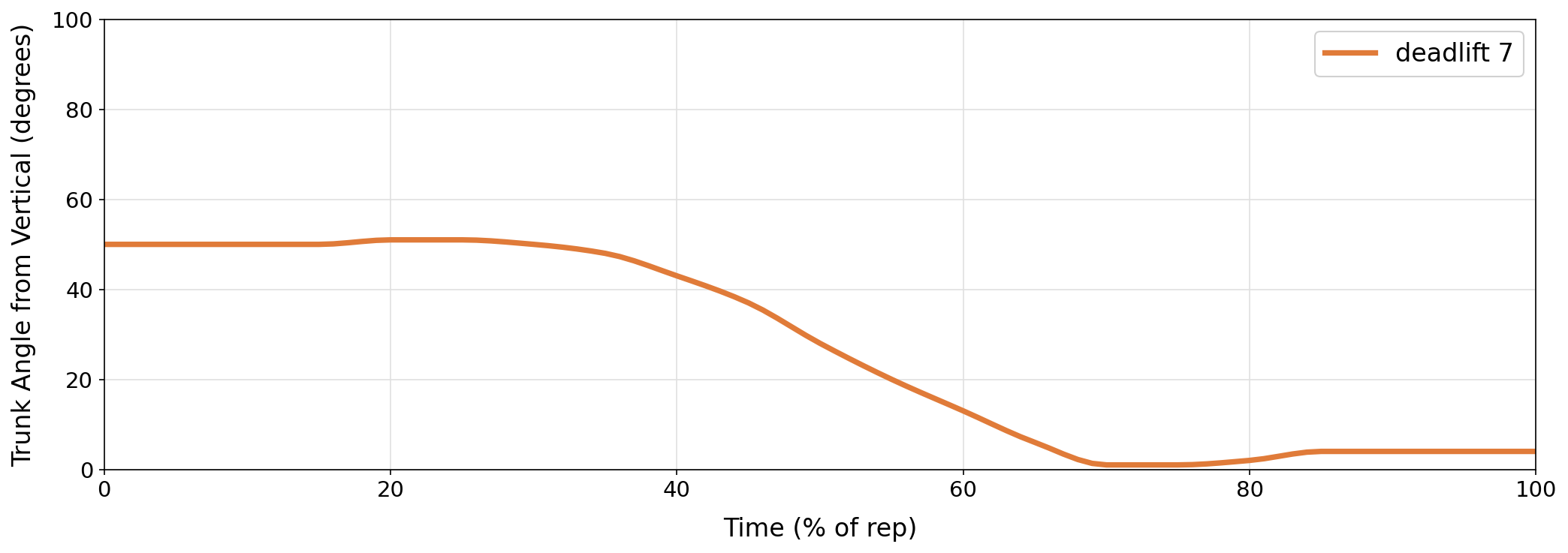}
        \caption{Time-normalised trunk angle from vertical.}
        \label{fig:deadlift_trunk}
    \end{subfigure}

    \caption{Joint-angle trajectories for a representative deadlift repetition.}
    \label{fig:deadlift}
\end{figure}

The hip-hinge angle increased smoothly throughout the movement, reflecting 
progressive hip extension from the bottom position to full lockout. The trunk 
angle began in a forward-leaning position and became more upright during ascent, indicating controlled trunk extension throughout the lift. Both signals exhibited a monotonic structure compared with the squat, reflecting the hinge-dominant nature of the movement. These results demonstrate that the framework can extract meaningful joint-angle trajectories beyond the squat, supporting its generalisability to compound movements with distinct biomechanical profiles.

\subsection{Limitations and Future Work}
The primary limitation of this framework is the sensitivity of 2D joint-angle estimation to camera orientation. As demonstrated in Section \ref{sec:camera sens}, even when the underlying posture remains unchanged, viewpoint changes introduce substantial distortion in estimated joint angles. Preliminary work into 3D landmark reconstruction suggests partial improvement under non-sagittal conditions but did not fully recover the expected range of motion, as the landmarks are inferred from a single view and remain subject to perspective ambiguity. Reliable deployment in uncontrolled environments will therefore require either stricter guidance on camera placement or more robust depth-estimation techniques.

The RMSE metric, whilst interpretable, weights all phases of the movement equally. Deviations at biomechanically critical points, such as the bottom position of the squat, may carry greater practical significance than deviations during transitional phases, and a phase-weighted metric could better reflect how technique is evaluated in practice.

The dataset used is relatively small and specific. Real-world deployment would involve a wider range of lighting conditions, camera placements, and lifter experience levels, which the current pipeline does not fully account for.
Expanding the dataset to include a wider range of lifters, experience levels, and recording environments would improve the external validity of the framework and support more robust evaluation of the similarity metric across populations.

\section{Conclusion}
This paper presents and evaluates a markerless pose estimation framework for assessing resistance-training technique from standard video footage, and 
successfully reconstructed joint-angle trajectories for the squat and deadlift, with mean usable frame rates of 99.7\% and 99.0\% respectively. A frame was considered usable when BlazePose successfully detected all landmarks required for the joint-angle calculation with valid coordinates. The usable-frame rate therefore represents the percentage of frames within a video that could be included in the biomechanical analysis. Squat trajectories demonstrated consistent biomechanical structure and strong agreement with a reference repetition under sagittal viewing conditions.
Intra-set analysis captured movement quality variations across a set, with early and late repetitions showing the greatest deviation from the reference repetition, consistent with movement initiation effects and fatigue.

Camera viewpoint emerged as the strongest practical constraint. Rotation experiments confirmed that 2D joint-angle estimates are highly sensitive to perspective distortion, and that non-sagittal recordings can produce misleading trajectories even when the underlying movement is technically sound. The extension of the pipeline to the deadlift produced meaningful hip-hinge and trunk trajectories consistent with expected biomechanics, supporting the generalisability of the approach to compound movements with distinct mechanical profiles. 

Overall, the results indicate that markerless pose estimation can provide accessible, interpretable biomechanical assessment outside laboratory environments. Whilst limitations around viewpoint sensitivity, occlusion, and metric design remain, the framework represents a practical step towards data-driven movement feedback for everyday training settings.

\section{Code Availability}
\label{sec:code}

The code used to implement the proposed framework will be released upon publication at
\href{https://github.com/ac2771/Markerless-Pose-Resistance-Training.git} {\texttt{github.com/ac2771/Markerless-Pose-Resistance-Training}}.

\begin{credits}
\subsubsection{\ackname}
NK was supported by the EPSRC LEAP Digital Health Hub grant EP/X031349/1.

\subsubsection{\discintname}
The authors have no competing interests to declare that are
relevant to the content of this article.
\end{credits}

\bibliographystyle{splncs04}
\bibliography{refs}

\end{document}